\documentclass[letterpaper]{article} % DO NOT CHANGE THIS
\usepackage{times}  % DO NOT CHANGE THIS
\usepackage{helvet}  % DO NOT CHANGE THIS
\usepackage{courier}  % DO NOT CHANGE THIS
\usepackage[hyphens]{url}  % DO NOT CHANGE THIS
\usepackage{graphicx} % DO NOT CHANGE THIS
\usepackage{natbib}  % DO NOT CHANGE THIS AND DO NOT ADD ANY OPTIONS TO IT
\usepackage{caption} % DO NOT CHANGE THIS AND DO NOT ADD ANY OPTIONS TO IT
\usepackage{algorithm}
\usepackage{algorithmic}
\usepackage{amsmath}
\usepackage{diagbox} 
\usepackage{xcolor}
\usepackage{multirow}
\usepackage{booktabs} % For \toprule, \m

\usepackage{bibentry}
\usepackage{amsmath} % For math expressions
\usepackage{authblk}
\usepackage{newfloat}
\usepackage{listings}
\begin{document}
\title{LE-PDE++: Mamba for accelerating PDEs Simulations}

\author[1,2]{Aoming Liang \thanks{Email: liangaoming@westlake.edu.cn}}
\author[3]{Zhaoyang Mu \thanks{Email: muzhaoyang@gmail.com}}
\author[2]{Qi Liu \thanks{Email: liuqi76@westlake.edu.cn}}
\author[2]{Ruipeng Li\thanks{Email: liruipeng@westlake.edu.cn, corresponding author}}
\author[2]{Mingming Ge\thanks{Email: gemingming@westlake.edu.cn, corresponding author}}
\author[2]{Dixia Fan\thanks{Email: fandixia@westlake.edu.cn, corresponding author}}

\affil[1]{Zhejiang University, Hangzhou, China}
\affil[2]{Westlake University, Hangzhou, China}
\affil[3]{Dalian Maritime University, Dalian, China}

\maketitle

% REMOVE THIS: bibentry
% This is only needed to show inline citations in the guidelines document. You should not need it and can safely delete it.

% END REMOVE bibentry

\begin{abstract}
Partial Differential Equations (PDEs) are foundational in modeling science and natural systems such as fluid dynamics and weather forecasting. The Latent Evolution of PDEs (LE-PDE) method is designed to address the computational intensity of classical and deep learning-based PDE solvers by proposing a scalable and efficient alternative. To enhance the efficiency and accuracy of LE-PDE, we incorporate the Mamba model(LE-PDE++)—an advanced machine learning model known for its predictive efficiency and robustness in handling complex dynamic systems with a progressive learning strategy. The LE-PDE++ was tested on several benchmark problems. The method demonstrated a marked reduction in computational time compared to traditional solvers and standalone deep learning models while maintaining high accuracy in predicting system behavior over time. Our method doubles the inference speed compared to the LE-PDE while retaining the same level of parameter efficiency, making it well-suited for scenarios requiring long-term predictions.

\end{abstract}


\section{Introduction}

Partial differential equations (PDEs) play a pivotal role in scientific and engineering fields, modeling the dynamic evolution of complex systems over time. These equations are indispensable for both forward prediction and reverse optimization, making them essential tools across various disciplines; these include weather forecasting \citep{holmstrom2016machine}, jet engine design \citep{yuksel2023review}, nuclear fusion \citep{pavone2023machine},  laser-plasma interaction \citep{dopp2023data}, and physical simulations \citep{wu2022learning}.

When tackling real-world challenges in science and engineering, the number of cells per time step can easily reach millions or more. This complexity presents a significant obstacle for traditional PDE solvers, which struggle to deliver rapid solutions at such scales. Moreover, inverse optimization tasks, such as inferring system parameters, encounter similar challenges, compounding the difficulties of modeling forward evolution \citep{biegler2003large}. In response to these limitations, numerous deep learning-based models have emerged, offering the potential to accelerate the solving of partial differential equations by orders of magnitude—often achieving speeds 10 to 1000 times faster by Fourier Neural Operator(FNO)\citep{li2020fourier}.

\textbf{Deep learning-based surrogate models:} In the research of surrogate models, there are three broad classes: pure data-driven, physical information-driven, and hybrid.
In the data-driven method, finite operator models that depend on grids have shown significant promise. \citep{raissi2018deep, rosofsky2023applications,sirignano2018dgm,guo2016convolutional,khoo2021solving}. infinite operator models show a good potential over geometry and sampling  \citep{li2020neural,li2020multipole,lu2021learning}.
For the physical information methods, researchers primarily incorporate information from known physical equations into the network through two main methods: hard constraints \citep{chalapathi2024scaling} and soft constraints\citep{kivcic2023adaptive}. These approaches ensure that the network adheres closely to established physical laws, enhancing the reliability and accuracy of the predictions.

However, the above methods still face the following challenges: 
\begin{enumerate}
    \item The models rely on an end-to-end mapping structure, and using CNNs as the base model leads to significant convolutional computation time, resulting in increased time complexity.
    \item The training mechanisms of these models are not yet well-defined, and no efficient learning approach enables the model to learn how to learn effectively.
\end{enumerate}

To address the first issue, we draw on the LE-PDE \citep{wu2022learning} and the Mamba\citep{gu2023mamba} model, utilizing latent space for rapid inference. This allows us to maintain several parameters while improving computational speed. We introduce a progressive learning mechanism to tackle the second issue, enabling the model to adapt and improve over time.
To differentiate from the LE-PDE, it will refer to the enhanced model as LE-PDE++ in the following discussion.

\begin{figure*}[h]
    \hspace{0.5cm}
    \includegraphics[width=1\textwidth]{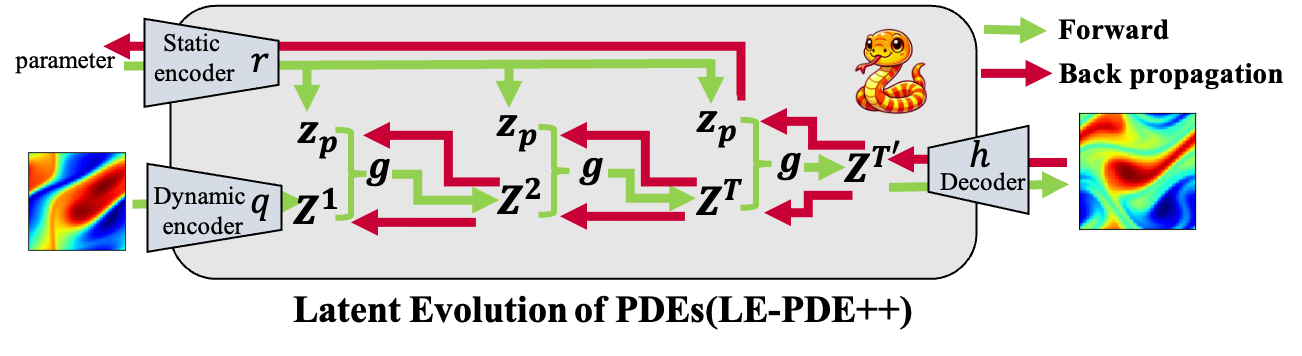}  
    \caption{The framework of LE-PDE++}
    \label{fig: show}
\end{figure*}

\section{Related Work}
In recent years, significant efforts have been devoted to addressing the challenges above.
\citet{calder2019pde} from the perspective of numerical analysis, accelerating the solution of PDE in momentum-based methods, such as Nesterov's accelerated gradient method and Polyak's heavy-ball method, can be interpreted through their variational formulations.  \citet{kuzmych2022accelerating} compared CNN-based methods and the finite element method, and experimental results demonstrate that CNNs can accelerate the solution process in regular domains. GNN-based models \citep{li2022graph} have been successfully applied in fluid-particle simulations. Neural Operators\citet{kovachki2023neural} learn a neural network (NN) that approximates a mapping between infinite-dimensional functions. While they have the advantage of being discretization invariant, they still require updating the state at each cell based on its neighboring cells (and potentially distant cells) given a specific discretization, which remains inefficient during inference time. \citet{wu2023learning} proposed a reinforcement learning-based controllable simulation method, which accelerates large-scale grid simulations. 

\citet{kumar2024synergistic} integrated a multi-task learning approach to tackle various PDE tasks. \citet{choi2024snn} describe a dynamic PDE by Hodge theory and proposed SNN-PDE to learn the physic system efficiently. Addressing the challenge of long-term stability in PDE solvers, \citet{wang2022respecting} introduced a novel framework for constructing neural PDE solvers that respect physical symmetries and conservation laws. \citet{xiong2024koopman} introduced a Koopman operator to learn nonlinear dynamics in the PDE solvers. \citet{brandstetter2022clifford} introduce a Clifford algebra computational layer, leveraging vector fields' rotational, translational, and projection properties to enable the model to learn more comprehensive physical representations. By combining data-driven learning with physics-based constraints, these approaches accelerate computation and enhance the accuracy and generalizability of PDE solvers across diverse scientific domains.

Upon on this work,  we conducted extensive experiments across diverse datasets to validate our contributions. Utilizing the Mamba model, we have significantly accelerated the inference process in the latent space. Moreover, we introduce a novel approach to progressive learning, aiming to enhance the adaptability and efficiency of the model. This approach optimizes performance and provides a flexible framework that can be tailored to various complex datasets, demonstrating substantial improvements over traditional methods.

\hspace{-2cm}
\section{Preliminaries} The LE-PDE model architecture comprises four key components in the Appendix \ref{LE-PDE details}:
\begin{align*}
q &: \text{dynamic encoder: } z^k = q(U^k) \\
r &: \text{static encoder: } z_p = r(p) \\
g &: \text{latent evolution model: } z^{k+1} = g(z^k, z_p) \\
h &: \text{decoder: } \hat{U}^{k+1} = h(z^{k+1})
\end{align*}

LE-PDE utilizes the temporal bundling technique (\citep{brandstetter2022message}) to enhance the representation of sequential data. This approach involves grouping input states $U^k$ across a fixed interval $S$ of consecutive time steps. Consequently, each latent vector $\mathbf{z}_k$ encodes these states bundle, and latent evolution predicts the next $\mathrm{z}_{k+1}$ for the subsequent $S$ steps. The parameter $S$, a hyperparameter, is adaptable to the specific problem, and setting $S=1$ results in no bundling. The autoregressive output $\hat{U}^{t+m}$ is defines as:
\begin{equation}
\begin{aligned}
\hat{U}^{t+m} &= h\left(z^{t+m}\right) \\
&\equiv h\left(g\left(\cdot, z_p\right)^{(m)} \circ z^t\right) \\
&\equiv h\left(g(\cdot, r(p))^{(m)} \circ q\left(U^t\right)\right)
\end{aligned}
\end{equation}

The training loss is as follows:
\begin{equation}
L = \frac{1}{K} \sum_{k=1}^K \left(L_{\text{multi-step}}^k + L_{\text{recons}}^k + L_{\text{consistency}}^k \right)
\end{equation}
Where
\begin{flalign*}
&L_{\text{multi-step}}^k = \sum_{m=1}^M \alpha_{m} \ell(\hat{U}^{k+m}, U^{k+m}), &\\
&L_{\text{recons}}^k = \ell(h(q(U^k)), U^k), &\\
&L_{\text{consistency}}^k = \sum_{m=1}^M \frac{\|g(\cdot, r(p))^{(m)} \circ q(U^k) - q(U^{k+m})\|_2^2}{\|q(U^{k+m})\|_2^2}. &
\end{flalign*}

However, existing LE-PDE suffer from \textbf{several drawbacks}:

\begin{enumerate}

    \item  \textbf{Slow in training and inference stage} Although Le-PDE effectively compresses information to the latent space by the dynamic encoder, the forward evolution in the latent space is progressive, making multi-step prediction significantly time-consuming.

    \item \textbf{Continuity conflict predication in the training objective} It is because the multi-step bundling strategy in the training objective function and the continuity of losses may conflict in highly non-linear problems, as the former loss term ensures the model's multi-step prediction capability, the latter requires that perpetuation in the latent space are not too large.

\end{enumerate}

\section{Our LE-PDE++ Framework}

This section provides a detailed explanation of the LE-PDE++ method and the progressive learning approach. Figure\ref{fig: show} outlines the model's architecture.

\subsection{Mamba for latent evolution}

\paragraph{LE-PDE++} utilizes the Mamba model in the latent space to transform the compressed latent vector. Specifically, the transformation is defined as follows:

\begin{equation}
\begin{aligned}
& h_t = \overline{\boldsymbol{A}} h_{t-1} + \overline{\boldsymbol{B}} x_t \\
& y_t = \mathbf{C} h_t \\
& \bar{K} = \left(C \overline{\boldsymbol{B}}, C \overline{\boldsymbol{A B}}, \ldots, C \overline{\boldsymbol{A}}^k \overline{\boldsymbol{B}}, \ldots\right) \\
& y = x * \bar{K}
\end{aligned}
\end{equation}

where $\overline{\mathbf{A}}$ is defined as $\exp(\Delta \mathbf{A})$, with $\Delta$ being a learnable sampling rate parameter, and $\overline{\mathbf{B}}$ is given by $(\Delta \mathbf{A})^{-1} (\exp(\Delta \mathbf{A}) - \mathbf{I}) \cdot \Delta \mathbf{B}$, where $\Delta$ also represents the learnable sampling rate parameter. $h_t$ is the latent vector in the time $t$. $\bar{K}$ is a causal convolution operator. $y_t$ is the output of Mamba.
We make the following assumptions here:
\begin{itemize}
    \item Linear time invariance: Most systems exhibit linear characteristics in a low-dimensional space, which was not present in the original LE-PDE model.
    \item Acceleration of model performance, which is significant for algorithms in autoregressive architectures
\end{itemize}

\subsection{Progressive sampling policy as the learning objective}

The idea of progressive sampling originates \citet{provost1999efficient,bengio2015scheduled,wang2021survey}, and this paper replaces long-term prediction sequences in the loss function by gradually adjusting the non-masked (i.e., model-visible) portion ratio.

1. **Linear Growth**:
\begin{equation}
r(n) = \min \left(1, \tau_0 + \left(1 - \tau_0\right) \frac{n}{N}\right),
\end{equation}

which represents a linear growth model. Additionally, we have also designed models with logarithmic and polynomial growth rates for comparison:

2. **Polynomial Growth**:
   \begin{equation}
   r(n) = \min \left(1, \tau_0 + \left(1 - \tau_0\right) \left(\frac{n}{N}\right)^p\right),
   \end{equation}
   Where \( p \) controls the polynomial growth rate.

3. **Logarithmic Growth**:
   \begin{equation}
   r(n) = \min \left(1, \tau_0 + \left(1 - \tau_0\right) \log\left(1 + \frac{n}{N}\right)\right).
   \end{equation}

\begin{figure}[h!] % 'h' 表示尽量将图像放在当前位置
    \centering % 图片居中
    \includegraphics[width=0.5\textwidth]{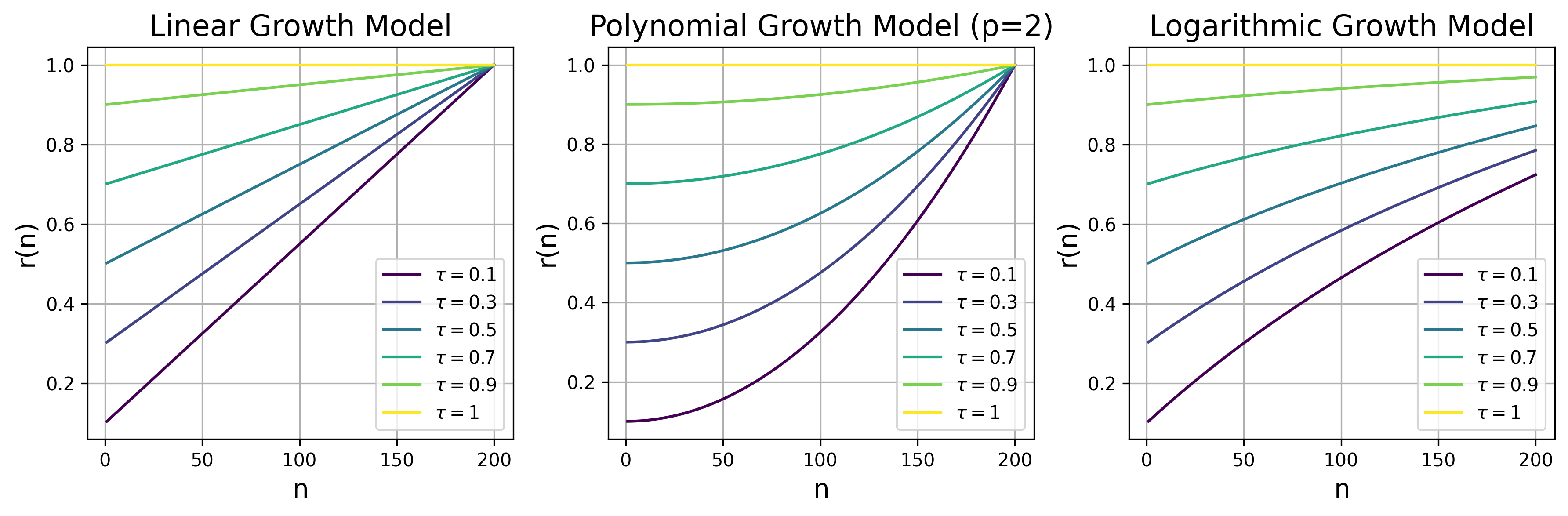} % 设定图片宽度为文本宽度的80%
    \caption{Progressive sampling ratio with three different Setting} % 图片标题
    \label{fig:example} % 标签，用于在文中引用
\end{figure}

Integrating this approach enables the model to initially predict with smaller time steps, as it is generally easier for neural networks to learn nearby dynamical behaviors. For more distant time steps, progressive sampling is used to maintain balance.
The advantage of using these functions over the traditional \(\alpha_{m}\) parameter in the LE-PDE is the simplicity of having a single hyperparameter to manage. This flexibility allows for more tailored encouragement of multi-step predictions, accommodating different growth rates as needed.

\section{Experiments}

In this study, we aim to compare and analyze the performance of LE-PDE++ and baseline methods in the domain of PDE simulations. We seek to address the following research questions:

\begin{enumerate}
    \item Does the Mamba model truly accelerate the forward inference in the latent space?
    \item What is the impact of different asymptotic sampling methods, and which asymptotic strategy is more universal?
    \item Can our approach surpass state-of-the-art (SOTA) methods while maintaining similar parameters?
\end{enumerate}

To address these questions, we will evaluate the models using two main aspects:
To address these questions, we will evaluate the models using two main aspects:
\begin{itemize}
    \item \textbf{Quality in Prediction Accuracy}: This is measured by single-step root mean squared error (RMSE) and global RMSE.
    \item \textbf{Inference time}: This is assessed by the speed of inference over global inference.
\end{itemize}

\subsection{Navier-Stokes Equation Datase (NS)}
The Navier-Stokes equations are fundamental in various scientific and engineering disciplines, encompassing applications such as weather forecasting, aerospace engineering, and hydrodynamics. The simulation of these equations becomes increasingly complex in the turbulent regime, characterized by intricate multiscale dynamics and chaotic behavior.

\begin{align}
\begin{aligned}
\partial_t w(t, x) + u(t, x) \cdot \nabla w(t, x) &= \nu \Delta w(t, x) + f(x) \\
\nabla \cdot u(t, x) &= 0 \\
w(0, x) &= w_0(x) \\
\end{aligned} 
\end{align}

Where, the vorticity is given by \( w(t, x) = \nabla \times u(t, x) \), the domain is discretized into a \(64 \times 64\) grid with a Reynolds number \(Re = 10^4\), indicating turbulence. $x \in (0,1)^2,  t \in (0,1]$.The dataset comprises 1200 trajectories, of which 1000 are used for training and 200 for testing.

\begin{figure}[h!]
    \centering
    \includegraphics[width=0.5\textwidth]{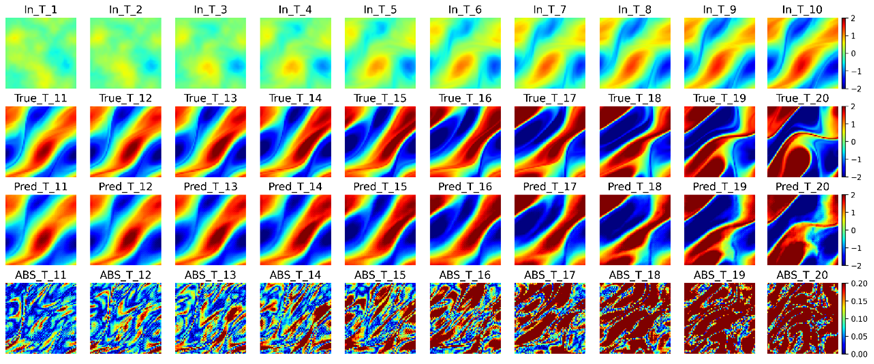}  % 
    \caption{Performace on NS of LE-PDE++, the first row is the input state, the second row is the ground truth, the third row is the output of LE-PDE++, the final row is an absolute error. }
    \label{fig:example}
\end{figure}

\subsection{Shallow Water Equation (SWE) }
The Shallow Water Equations describe phenomena related to large-scale ocean wave motions.
\begin{gather}
\begin{aligned}
\frac{\partial u(x, y, t)}{\partial t} &= -g \frac{\partial \eta(x, y, t)}{\partial x} \\
\frac{\partial v(x, y, t)}{\partial t} &= -g \frac{\partial \eta(x, y, t)}{\partial y} \\
\frac{\partial u(x, y, t)}{\partial x} + \frac{\partial v(x, y, t)}{\partial y} &= -\frac{1}{H} \frac{\partial \eta(x, y, t)}{\partial t}
\end{aligned}
\end{gather}
Where \( u(x, y, t) \) represents the horizontal velocity component of the fluid at position \((x, y)\) and time \(t\), \( v(x, y, t) \) is the vertical velocity component of the fluid at the same position and time, and \( \eta(x, y, t) \) denotes the surface elevation of the fluid at position \((x, y)\) and time \(t\). The parameter \( g=1 \) represents the dimensionless gravitational acceleration, and \( H =100\) is the water depth. $x \in (0,1)^2,  t \in (0,1]$, the spatial resolution is \( 128 \times 128 \).

In this study, we solve these equations for 100 trajectories, where 70 are used for training and 30 for testing.

\begin{figure}[h]
    \centering
    \includegraphics[width=0.5\textwidth]{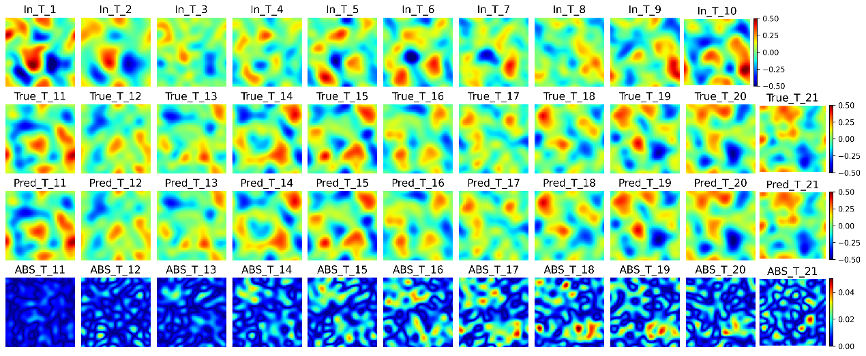}  % 
    \caption{Performace on SWE of LE-PDE++ }
    \label{fig:fig_swe}
\end{figure}
\

\subsection{Pollutant Transport Equation (PTE) }
 The following diffusion equation governs the pollutant dispersion:

\begin{equation}
\frac{\partial C(x,t)}{\partial t} = D \nabla^2 C(x,t) + S(x,t)
\end{equation}

Where \( C(x, t) \) represents the pollutant concentration at position \( x \) and time \( t \), \( D \) is the diffusion coefficient, \( \nabla^2 \) is the Laplace operator, and \( S(x, t) \) is the source term. The numerical simulations are performed on a grid with dimensions \( 512 \times 512 \), representing a spatial domain of \( 3 \text{ km} \times 3 \text{ km} \). This grid resolution allows for detailed modeling of the pollutant dispersion process. The time step used in the simulation is \( 5 \text{ s} \), and we select \( 21 \) time steps for our analysis. The data set utilized for the simulations is sourced from Alibaba's Tianchi.

The dataset comprises 121 simulated cases, each with randomly placed pollution sources and four wind directions. We randomly select 100 cases for the training set and 21 cases for the testing set.

\begin{figure}[h!]
    \centering
    \includegraphics[width=0.5\textwidth]{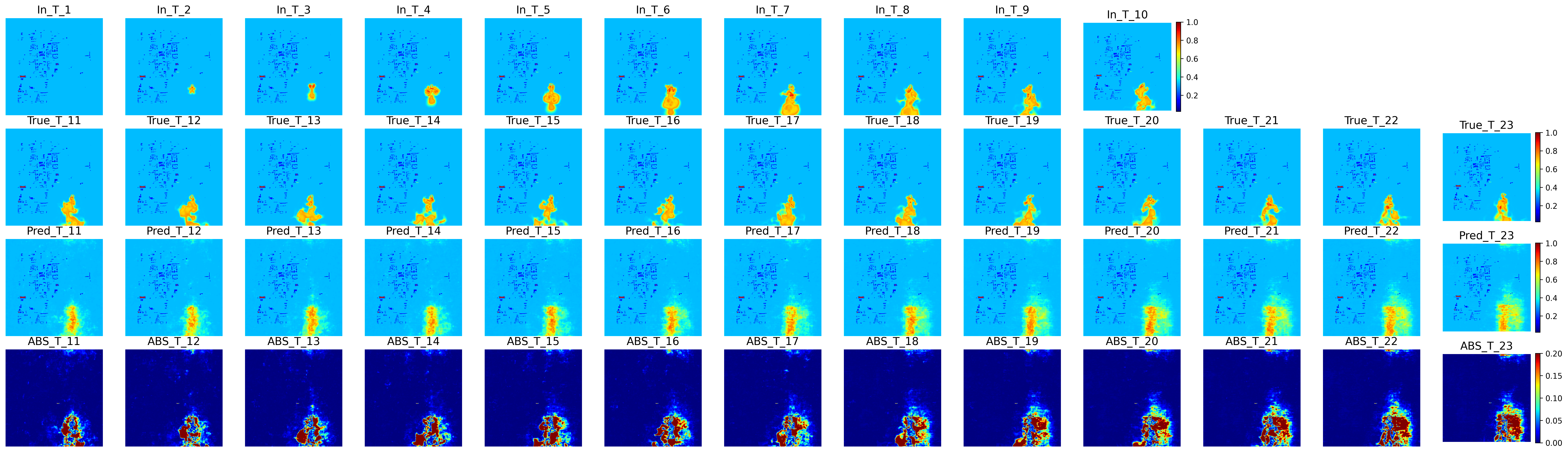}  % 
    \caption{Performace on PTE of LE-PDE++ }
    \label{fig:fig_pte}
\end{figure}

\subsection{Ablation and Comparison Results}
Ablation experiments primarily measure whether adding the Mamba model accelerates the processing time of the LE-PDE method. Comparative experiments demonstrate that our approach offers certain advantages over four baseline methods: FNO, WNO\citep{navaneeth2024physics}, UNO\citep{azizzadenesheli2024neural}, and the original LE-PDE. The specific settings of these baseline methods are detailed in the appendix.  The following experiments were conducted on NVIDIA TESLA V100.

\begin{table}[h!]
    \centering
    \caption{Ablation Study about LE-PDE++ and LE-PDE on NS}
    \begin{tabular}{c|c|c}
        \toprule
    Metric & LE-PDE & LE-PDE++$^{*}$\\
        \midrule
        Inference(ms) & 12$\pm$2.5 & 6$\pm$0.6\\
         \midrule
        Parameters($10^{7}$)& 3.8 & 3.2 \\
         \midrule
        RMSE& 0.26& 0.31\\
        \bottomrule
    \end{tabular}
    \label{tab:ablation}
   
\end{table}

From Table\ref{tab:ablation}, the LE-PDE++ model, which replaces the evolution model with Mamba, shows a twofold increase in inference speed. However, the RMSE reaches 0.31. Note that in this experiment, progressive sampling was not used. We analyzed the issue and concluded that while the model's linear inference speed is very fast, it performs poorly on nonlinear datasets. To address this issue, we need to introduce progressive sampling, gradually allowing the model to learn to predict over longer time horizons.

To address the issue of not using progressive sampling, we designed several experiments with initial values of 0.1, 0.3, 0.5, 0.7, 0.9, and 1, each corresponding to three different progressive strategies. It is worth noting that in practical applications, $r(n)$ is set to an integer.
\begin{table}[h!]
    \centering
    \caption{Progressive Sampling ratio and RMSE performance in the LE-PDE++ on NS}

    \begin{tabular}{cccc}
        \toprule
        \diagbox{$\tau_{0}$}{Policy} & Linear & Log& Poly \\
        \midrule
        $\tau_{0}=0.1$ & 0.24  & 0.23 & 0.28\\
        \midrule
       $\tau_{0}=0.3$ & 0.27 & \textbf{0.21} & 0.35\\
        \midrule
       $\tau_{0}=0.5$ & 0.25 &  0.29& 0.24 \\
        \midrule
       $\tau_{0}=0.7$  & 0.30 &  0.24 & 0.29  \\
        \midrule
        $\tau_{0}=0.9$  & 0.30 &  0.30 & 0.29 \\
        \midrule
        $\tau_{0}=1$  & 0.31 & 0.31& 0.31 \\
        \bottomrule
    \end{tabular}
    \label{tab:progressive}
\end{table}

The table \ref{tab:progressive} shows that the choice of $\tau_{0}$ is crucial for the results of LE-PDE++. Among the three progressive strategies, the Mamba model performs better with Logarithmic growth than other types, and setting a smaller value tends to yield better results. In the following experiments, we consistently employed this Logarithmic policy for training.

In the final experiment, we aim to compare the current baseline models horizontally. Details of their configurations are provided in the appendix. The basic principle was to keep their parameter levels comparable and use autoregressive methods to predict the corresponding solutions.

\begin{table}[h!]

\caption{Comparison on the Baseline methods overall the dataset}
\hspace{-1cm}   % 这是正确的语法
\begin{tabular}{c|c|c|c|c}

\toprule
Dataset&Baselines & Parameter$(10^{7})$ & RMSE & Inference(ms) \\
\midrule
\multirow{4}{*}{NS} & LE-PDE++ & 3.2&0.21$^{*}$&6$^{*}$ \\
\cline{2-5}
& FNO &3.4 &0.22& 7.5\\
\cline{2-5}
& UNO &4.5 & 0.24& 12\\
\cline{2-5}
& WNO &4.2 &0.28 & 21\\
\midrule
\multirow{4}{*}{SWE} & LE-PDE++ & 6.2 & 0.30& 9$^{*}$ \\
\cline{2-5}
& FNO & 5.8&0.25 & 14\\
\cline{2-5}
& UNO & 7.8&0.28 & 30\\
\cline{2-5}
& WNO &8.2 & 0.30& 45\\
\midrule
\multirow{4}{*}{PTE} & LE-PDE++ &12.4 & 0.42$^{*}$ & 24$^{*}$ \\
\cline{2-5}
& FNO &14.8 &0.47 & 80 \\
\cline{2-5}
& UNO &18.4 &0.44 &120 \\
\cline{2-5}
& WNO & 20.2& 0.48& 300\\
\bottomrule
\end{tabular}
\label{tab:comparison}
\end{table}

As shown in the table \ref{tab:comparison}, the results for LE-PDE++ (marked with an asterisk) are generally on par with other models while achieving competitive accuracy. The ablation study highlights that progressive sampling methods are crucial for ensuring accurate long-term predictions. We anticipate that our approach will significantly advance the acceleration of simulations for PDEs, which are essential in science and engineering. The inference speed of LE-PDE++ is significantly faster compared to other models. This advantage is attributed to the benefits of linear models. As the grid size increases, the dynamic encoder compresses the model and enables rapid inference by leveraging the latent space.

\section{Conclusion}

In this work, we have introduced the LE-PDE++ framework, a novel surrogate modeling approach that replaces the traditional evolution model with the Mamba model. This enhancement has effectively doubled the inference speed compared to the original LE-PDE. Furthermore, we have introduced the concept of progressive sampling, which enables the model to extend its predictive capabilities over longer dynamic behaviors—marking a pioneering attempt in surrogate modeling. Future work will explore and address the inherent uncertainty aspects of the model. Additionally, our method has demonstrated a significant improvement in performance on complex datasets such as PTE, achieving speeds 4 to 15 times faster than baseline methods.

% \section{Acknowledgments}

% This work was supported by the Priority Postdoctoral Projects in Zhejiang Province (ZJ2023023), GuangDong Basic and Applied Basic Research Foundation (2024A1515010711). The author sincerely thanks the High-Performance Computing Center at Westlake University for their support.

\bibliographystyle{plainnat}

\bibliography{aaai25}

\begin{thebibliography}{34}
\providecommand{\natexlab}[1]{#1}
\providecommand{\url}[1]{\texttt{#1}}
\expandafter\ifx\csname urlstyle\endcsname\relax
  \providecommand{\doi}[1]{doi: #1}\else
  \providecommand{\doi}{doi: \begingroup \urlstyle{rm}\Url}\fi

\bibitem[Azizzadenesheli et~al.(2024)Azizzadenesheli, Kovachki, Li, Liu-Schiaffini, Kossaifi, and Anandkumar]{azizzadenesheli2024neural}
Kamyar Azizzadenesheli, Nikola Kovachki, Zongyi Li, Miguel Liu-Schiaffini, Jean Kossaifi, and Anima Anandkumar.
\newblock Neural operators for accelerating scientific simulations and design.
\newblock \emph{Nature Reviews Physics}, pages 1--9, 2024.

\bibitem[Bengio et~al.(2015)Bengio, Vinyals, Jaitly, and Shazeer]{bengio2015scheduled}
Samy Bengio, Oriol Vinyals, Navdeep Jaitly, and Noam Shazeer.
\newblock Scheduled sampling for sequence prediction with recurrent neural networks.
\newblock \emph{Advances in neural information processing systems}, 28, 2015.

\bibitem[Biegler et~al.(2003)Biegler, Ghattas, Heinkenschloss, and van Bloemen~Waanders]{biegler2003large}
Lorenz~T Biegler, Omar Ghattas, Matthias Heinkenschloss, and Bart van Bloemen~Waanders.
\newblock Large-scale pde-constrained optimization: an introduction.
\newblock In \emph{Large-scale PDE-constrained optimization}, pages 3--13. Springer, 2003.

\bibitem[Brandstetter et~al.(2022{\natexlab{a}})Brandstetter, Berg, Welling, and Gupta]{brandstetter2022clifford}
Johannes Brandstetter, Rianne van~den Berg, Max Welling, and Jayesh~K Gupta.
\newblock Clifford neural layers for pde modeling.
\newblock \emph{arXiv preprint arXiv:2209.04934}, 2022{\natexlab{a}}.

\bibitem[Brandstetter et~al.(2022{\natexlab{b}})Brandstetter, Worrall, and Welling]{brandstetter2022message}
Johannes Brandstetter, Daniel Worrall, and Max Welling.
\newblock Message passing neural pde solvers.
\newblock \emph{arXiv preprint arXiv:2202.03376}, 2022{\natexlab{b}}.

\bibitem[Calder and Yezzi(2019)]{calder2019pde}
Jeff Calder and Anthony Yezzi.
\newblock Pde acceleration: a convergence rate analysis and applications to obstacle problems.
\newblock \emph{Research in the Mathematical Sciences}, 6\penalty0 (4):\penalty0 35, 2019.

\bibitem[Chalapathi et~al.(2024)Chalapathi, Du, and Krishnapriyan]{chalapathi2024scaling}
Nithin Chalapathi, Yiheng Du, and Aditi Krishnapriyan.
\newblock Scaling physics-informed hard constraints with mixture-of-experts.
\newblock \emph{arXiv preprint arXiv:2402.13412}, 2024.

\bibitem[Choi et~al.(2024)Choi, Chen, Lee, Kim, and Gel]{choi2024snn}
Jae Choi, Yuzhou Chen, Huikyo Lee, Hyun Kim, and Yulia~R Gel.
\newblock Snn-pde: Learning dynamic pdes from data with simplicial neural networks.
\newblock In \emph{Proceedings of the AAAI Conference on Artificial Intelligence}, volume~38, pages 11561--11569, 2024.

\bibitem[D{\"o}pp et~al.(2023)D{\"o}pp, Eberle, Howard, Irshad, Lin, and Streeter]{dopp2023data}
Andreas D{\"o}pp, Christoph Eberle, Sunny Howard, Faran Irshad, Jinpu Lin, and Matthew Streeter.
\newblock Data-driven science and machine learning methods in laser--plasma physics.
\newblock \emph{High Power Laser Science and Engineering}, 11:\penalty0 e55, 2023.

\bibitem[Gu and Dao(2023)]{gu2023mamba}
Albert Gu and Tri Dao.
\newblock Mamba: Linear-time sequence modeling with selective state spaces.
\newblock \emph{arXiv preprint arXiv:2312.00752}, 2023.

\bibitem[Guo et~al.(2016)Guo, Li, and Iorio]{guo2016convolutional}
Xiaoxiao Guo, Wei Li, and Francesco Iorio.
\newblock Convolutional neural networks for steady flow approximation.
\newblock In \emph{Proceedings of the 22nd ACM SIGKDD international conference on knowledge discovery and data mining}, pages 481--490, 2016.

\bibitem[Holmstrom et~al.(2016)Holmstrom, Liu, and Vo]{holmstrom2016machine}
Mark Holmstrom, Dylan Liu, and Christopher Vo.
\newblock Machine learning applied to weather forecasting.
\newblock \emph{Meteorol. Appl}, 10\penalty0 (1):\penalty0 1--5, 2016.

\bibitem[Khoo et~al.(2021)Khoo, Lu, and Ying]{khoo2021solving}
Yuehaw Khoo, Jianfeng Lu, and Lexing Ying.
\newblock Solving parametric pde problems with artificial neural networks.
\newblock \emph{European Journal of Applied Mathematics}, 32\penalty0 (3):\penalty0 421--435, 2021.

\bibitem[Ki{\v{c}}i{\'c} et~al.(2023)Ki{\v{c}}i{\'c}, Vlachas, Arampatzis, Chatzimanolakis, Guibas, and Koumoutsakos]{kivcic2023adaptive}
Ivica Ki{\v{c}}i{\'c}, Pantelis~R Vlachas, Georgios Arampatzis, Michail Chatzimanolakis, Leonidas Guibas, and Petros Koumoutsakos.
\newblock Adaptive learning of effective dynamics for online modeling of complex systems.
\newblock \emph{Computer Methods in Applied Mechanics and Engineering}, 415:\penalty0 116204, 2023.

\bibitem[Kovachki et~al.(2023)Kovachki, Li, Liu, Azizzadenesheli, Bhattacharya, Stuart, and Anandkumar]{kovachki2023neural}
Nikola Kovachki, Zongyi Li, Burigede Liu, Kamyar Azizzadenesheli, Kaushik Bhattacharya, Andrew Stuart, and Anima Anandkumar.
\newblock Neural operator: Learning maps between function spaces with applications to pdes.
\newblock \emph{Journal of Machine Learning Research}, 24\penalty0 (89):\penalty0 1--97, 2023.

\bibitem[Kumar et~al.(2024)Kumar, Goswami, Kontolati, Shields, and Karniadakis]{kumar2024synergistic}
Varun Kumar, Somdatta Goswami, Katiana Kontolati, Michael~D Shields, and George~Em Karniadakis.
\newblock Synergistic learning with multi-task deeponet for efficient pde problem solving.
\newblock \emph{arXiv preprint arXiv:2408.02198}, 2024.

\bibitem[Kuzmych and Novotarskyi(2022)]{kuzmych2022accelerating}
Valentyn Kuzmych and Mykhailo Novotarskyi.
\newblock Accelerating simulation of the pde solution by the structure of the convolutional neural network modifying.
\newblock In \emph{The International Conference on Artificial Intelligence and Logistics Engineering}, pages 3--15. Springer, 2022.

\bibitem[Li and Farimani(2022)]{li2022graph}
Zijie Li and Amir~Barati Farimani.
\newblock Graph neural network-accelerated lagrangian fluid simulation.
\newblock \emph{Computers \& Graphics}, 103:\penalty0 201--211, 2022.

\bibitem[Li et~al.(2020{\natexlab{a}})Li, Kovachki, Azizzadenesheli, Liu, Bhattacharya, Stuart, and Anandkumar]{li2020fourier}
Zongyi Li, Nikola Kovachki, Kamyar Azizzadenesheli, Burigede Liu, Kaushik Bhattacharya, Andrew Stuart, and Anima Anandkumar.
\newblock Fourier neural operator for parametric partial differential equations.
\newblock \emph{arXiv preprint arXiv:2010.08895}, 2020{\natexlab{a}}.

\bibitem[Li et~al.(2020{\natexlab{b}})Li, Kovachki, Azizzadenesheli, Liu, Bhattacharya, Stuart, and Anandkumar]{li2020neural}
Zongyi Li, Nikola Kovachki, Kamyar Azizzadenesheli, Burigede Liu, Kaushik Bhattacharya, Andrew Stuart, and Anima Anandkumar.
\newblock Neural operator: Graph kernel network for partial differential equations.
\newblock \emph{arXiv preprint arXiv:2003.03485}, 2020{\natexlab{b}}.

\bibitem[Li et~al.(2020{\natexlab{c}})Li, Kovachki, Azizzadenesheli, Liu, Stuart, Bhattacharya, and Anandkumar]{li2020multipole}
Zongyi Li, Nikola Kovachki, Kamyar Azizzadenesheli, Burigede Liu, Andrew Stuart, Kaushik Bhattacharya, and Anima Anandkumar.
\newblock Multipole graph neural operator for parametric partial differential equations.
\newblock \emph{Advances in Neural Information Processing Systems}, 33:\penalty0 6755--6766, 2020{\natexlab{c}}.

\bibitem[Lu et~al.(2021)Lu, Jin, Pang, Zhang, and Karniadakis]{lu2021learning}
Lu~Lu, Pengzhan Jin, Guofei Pang, Zhongqiang Zhang, and George~Em Karniadakis.
\newblock Learning nonlinear operators via deeponet based on the universal approximation theorem of operators.
\newblock \emph{Nature machine intelligence}, 3\penalty0 (3):\penalty0 218--229, 2021.

\bibitem[Navaneeth et~al.(2024)Navaneeth, Tripura, and Chakraborty]{navaneeth2024physics}
N~Navaneeth, Tapas Tripura, and Souvik Chakraborty.
\newblock Physics informed wno.
\newblock \emph{Computer Methods in Applied Mechanics and Engineering}, 418:\penalty0 116546, 2024.

\bibitem[Pavone et~al.(2023)Pavone, Merlo, Kwak, and Svensson]{pavone2023machine}
A~Pavone, A~Merlo, S~Kwak, and J~Svensson.
\newblock Machine learning and bayesian inference in nuclear fusion research: an overview.
\newblock \emph{Plasma Physics and Controlled Fusion}, 65\penalty0 (5):\penalty0 053001, 2023.

\bibitem[Provost et~al.(1999)Provost, Jensen, and Oates]{provost1999efficient}
Foster Provost, David Jensen, and Tim Oates.
\newblock Efficient progressive sampling.
\newblock In \emph{Proceedings of the fifth ACM SIGKDD international conference on Knowledge discovery and data mining}, pages 23--32, 1999.

\bibitem[Raissi(2018)]{raissi2018deep}
Maziar Raissi.
\newblock Deep hidden physics models: Deep learning of nonlinear partial differential equations.
\newblock \emph{Journal of Machine Learning Research}, 19\penalty0 (25):\penalty0 1--24, 2018.

\bibitem[Rosofsky et~al.(2023)Rosofsky, Al~Majed, and Huerta]{rosofsky2023applications}
Shawn~G Rosofsky, Hani Al~Majed, and EA~Huerta.
\newblock Applications of physics informed neural operators.
\newblock \emph{Machine Learning: Science and Technology}, 4\penalty0 (2):\penalty0 025022, 2023.

\bibitem[Sirignano and Spiliopoulos(2018)]{sirignano2018dgm}
Justin Sirignano and Konstantinos Spiliopoulos.
\newblock Dgm: A deep learning algorithm for solving partial differential equations.
\newblock \emph{Journal of computational physics}, 375:\penalty0 1339--1364, 2018.

\bibitem[Wang et~al.(2022)Wang, Sankaran, and Perdikaris]{wang2022respecting}
Sifan Wang, Shyam Sankaran, and Paris Perdikaris.
\newblock Respecting causality is all you need for training physics-informed neural networks.
\newblock \emph{arXiv preprint arXiv:2203.07404}, 2022.

\bibitem[Wang et~al.(2021)Wang, Chen, and Zhu]{wang2021survey}
Xin Wang, Yudong Chen, and Wenwu Zhu.
\newblock A survey on curriculum learning.
\newblock \emph{IEEE transactions on pattern analysis and machine intelligence}, 44\penalty0 (9):\penalty0 4555--4576, 2021.

\bibitem[Wu et~al.(2022)Wu, Maruyama, and Leskovec]{wu2022learning}
Tailin Wu, Takashi Maruyama, and Jure Leskovec.
\newblock Learning to accelerate partial differential equations via latent global evolution.
\newblock \emph{Advances in Neural Information Processing Systems}, 35:\penalty0 2240--2253, 2022.

\bibitem[Wu et~al.(2023)Wu, Maruyama, Zhao, Wetzstein, and Leskovec]{wu2023learning}
Tailin Wu, Takashi Maruyama, Qingqing Zhao, Gordon Wetzstein, and Jure Leskovec.
\newblock Learning controllable adaptive simulation for multi-resolution physics.
\newblock \emph{arXiv preprint arXiv:2305.01122}, 2023.

\bibitem[Xiong et~al.(2024)Xiong, Huang, Zhang, Deng, Sun, and Tian]{xiong2024koopman}
Wei Xiong, Xiaomeng Huang, Ziyang Zhang, Ruixuan Deng, Pei Sun, and Yang Tian.
\newblock Koopman neural operator as a mesh-free solver of non-linear partial differential equations.
\newblock \emph{Journal of Computational Physics}, page 113194, 2024.

\bibitem[Y{\"u}ksel et~al.(2023)Y{\"u}ksel, B{\"o}rkl{\"u}, Sezer, and Canyurt]{yuksel2023review}
Nurullah Y{\"u}ksel, H{\"u}seyin~R{\i}za B{\"o}rkl{\"u}, H{\"u}seyin~K{\"u}r{\c{s}}ad Sezer, and Olcay~Ersel Canyurt.
\newblock Review of artificial intelligence applications in engineering design perspective.
\newblock \emph{Engineering Applications of Artificial Intelligence}, 118:\penalty0 105697, 2023.

\end{thebibliography}

\end{document}